\documentclass{ecai} 
\usepackage{latexsym}
\usepackage{amssymb}
\usepackage{amsmath}
\usepackage{amsthm}
\usepackage{booktabs}
\usepackage{enumitem}
\usepackage{graphicx}
\usepackage{color}
\usepackage{hyperref}
\usepackage{multirow}  % 确保包含这个包
\usepackage{natbib}
\usepackage[belowskip=10pt]{caption}
\newcommand{\BibTeX}{B\kern-.05em{\sc i\kern-.025em b}\kern-.08em\TeX}

\begin{document}

%%%%%%%%%%%%%%%%%%%%%%%%%%%%%%%%%%%%%%%%%%%%%%%%%%%%%%%%%%%%%%%%%%%%%%%%

\begin{frontmatter}

%%% Use this command to specify your submission number.
%%% In doubleblind mode, it will be printed on the first page.

% \paperid{7636} 

%%% Use this command to specify the title of your paper.

\title{LAMM-ViT: AI Face Detection via Layer-Aware Modulation of Region-Guided Attention}

%%% Use this combinations of commands to specify all authors of your 
%%% paper. Use \fnms{} and \snm{} to indicate everyone's first names 
%%% and surname. This will help the publisher with indexing the 
%%% proceedings. Please use a reasonable approximation in case your 
%%% name does not neatly split into "first names" and "surname".
%%% Specifying your ORCID digital identifier is optional. 
%%% Use the \thanks{} command to indicate one or more corresponding 
%%% authors and their email address(es). If so desired, you can specify
%%% author contributions using the \footnote{} command.

% \author[A]{\fnms{First}~\snm{Author}\orcid{....-....-....-....}\thanks{Corresponding Author. Email: somename@university.edu.}\footnote{Equal contribution.}}
% \author[B]{\fnms{Second}~\snm{Author}\orcid{....-....-....-....}\footnotemark}
% \author[B,C]{\fnms{Third}~\snm{Author}\orcid{....-....-....-....}} 

% \address[A]{Short Affiliation of First Author}
% \address[B]{Short Affiliation of Second Author and Third Author}
% \address[C]{Short Alternate Affiliation of Third Author}

\author[A]{\fnms{Jiangling}~\snm{Zhang}}
\author[A]{\fnms{Weijie}~\snm{Zhu}}
\author[A]{\fnms{Jirui}~\snm{Huang}}
\author[A]{\fnms{Yaxiong}~\snm{Chen}\thanks{Corresponding Author. Email: yaxiong.chen@whut.edu.cn}}

\address[A]{Wuhan University of Technology}

%%% Use this environment to include an abstract of your paper.

\begin{abstract}
Detecting AI-synthetic faces presents a critical challenge: it is hard to capture consistent structural relationships between facial regions across diverse generation techniques. Current methods, which focus on specific artifacts rather than fundamental inconsistencies, often fail when confronted with novel generative models. To address this limitation, we introduce Layer-aware Mask Modulation Vision Transformer (LAMM-ViT), a Vision Transformer designed for robust facial forgery detection. This model integrates distinct Region-Guided Multi-Head Attention (RG-MHA) and Layer-aware Mask Modulation (LAMM) components within each layer. RG-MHA utilizes facial landmarks to create regional attention masks, guiding the model to scrutinize architectural inconsistencies across different facial areas. Crucially, the separate LAMM module dynamically generates layer-specific parameters, including mask weights and gating values, based on network context. These parameters then modulate the behavior of RG-MHA, enabling adaptive adjustment of regional focus across network depths. This architecture facilitates the capture of subtle, hierarchical forgery cues ubiquitous among diverse generation techniques, such as GANs and Diffusion Models. In cross-model generalization tests, LAMM-ViT demonstrates superior performance, achieving 94.09\% mean ACC (a +5.45\% improvement over SoTA) and 98.62\% mean AP (a +3.09\% improvement). These results demonstrate LAMM-ViT's exceptional ability to generalize and its potential for reliable deployment against evolving synthetic media threats.

\end{abstract}

\end{frontmatter}

\section{Introduction}

Recent advancements in generative models, particularly Generative Adversarial Networks (GANs) \citep{goodfellow2014generative,  karras2019style} and Diffusion Models (DMs) \citep{dhariwal2021diffusion, rombach2022high}, have revolutionized the creation of synthetic facial images. These models now generate faces that are virtually indistinguishable from authentic photographs, achieving unprecedented levels of photorealism. While this technology offers legitimate applications in entertainment and privacy protection, it also raises significant concerns about potential abuse to create fake profiles, spread misinformation and undermine public trust in visual media\citep{chen2022ost, wang2020cnn}. As highlighted by Liu et al. \citep{liu2021spatial}, the realistic nature of synthetic human face images generated by models like StyleGAN and diffusion-based methods poses serious social trust concerns due to their potential exploitation for malicious purposes.

\begin{figure}[t]
    \centering
    \includegraphics[width=\columnwidth]{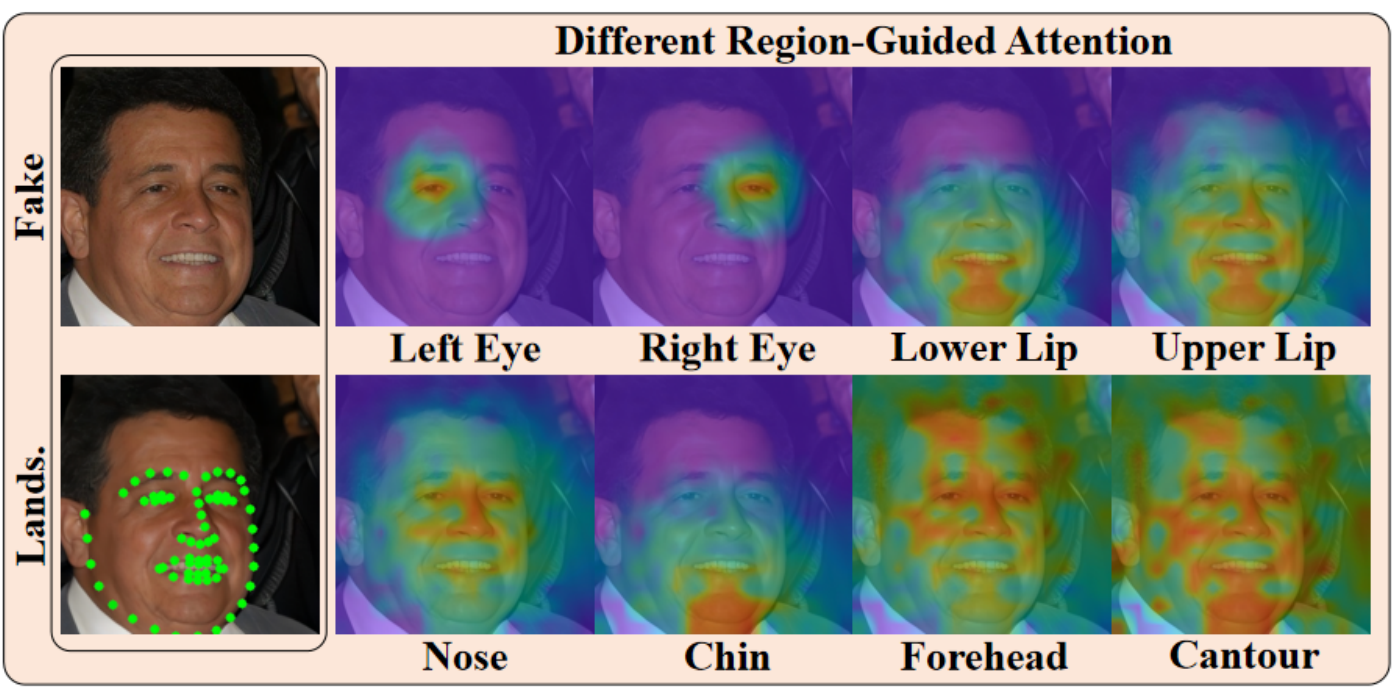}
    \caption{Example visualization of region-guided attention patterns generated by our LAMM-ViT model. The visualization demonstrates how different our region-guided attention heads focus on distinct facial regions with minimal overlap when analyzing AI-synthetic faces. Our method captures diverse forgery clues across various generative techniques, including subtle inconsistencies in texture patterns, unnatural symmetry, blending artifacts, and structural irregularities that persist across different generation methods.}
    \label{fig:features}
\end{figure}

Despite significant research efforts in synthetic image detection, most existing approaches face a critical limitation: poor generalization to new generation techniques not seen during training \citep{wang2020cnn}. This challenge stems from the fact that different generative models introduce different artifacts and patterns in their output. As observed by Wang et al. \citep{wang2023dire}, the generation processes between different models (e.g., GANs, VAEs, diffusion models) are entirely different, rendering previously developed detectors ineffective when confronted with images from novel generation methods. Furthermore, Jeong et al. \citep{jeong2022frepgan} note that owing to extraordinary advancements in synthesis technology, an increasing array of distinctive frequency-level artifact representations have emerged, further complicating detection.

Current detection methods fall into two categories: space-based methods analyzing pixel-level patterns and frequency-based methods examining spectral properties. Spatial-based approaches often employ CNN classifiers with various data pre-processing or augmentation strategies \citep{liu2021spatial, wang2020cnn}, while others target specific fingerprints left by the generation techniques \citep{ojha2023towards}. However, Wang et al. \citep{wang2023dire} note that classifiers trained on certain generators (like ProGAN) struggle with fake images from unfamiliar sources (such as diffusion models).  Frequency-domain methods \citep{durall2020watch, wang2020cnn} exploit abnormalities in the spectrum of synthetic images, particularly those caused by upsampling operations in generation pipelines. These approaches show promise but often fail against newer generation techniques, producing fewer detectable artifacts \citep{tan2024frequency}.

In this paper, we propose a novel detection approach that exploits a common vulnerability across diverse generation models: their inability to maintain consistent relationships between facial structures. Our main insight is that while modern generative models are good at creating globally coherent faces, they often introduce subtle inconsistencies in the relationships between facial regions that can be detected by paying careful attention to these regions. This view is consistent with the observation by He et al. that self-supervised models that examine global structure provide a more comprehensive perspective for detecting synthetic content.

To leverage this insight, we present a Mask-Guided Vision Transformer architecture with Layer-aware Mask Modulation (LAMM), which dynamically focuses on critical facial regions and their interrelationships at various feature abstraction levels. Unlike previous approaches that rely solely on spatial or frequency domain analysis, our method uses facial landmarks to create region-specific attention masks that guide the model toward discriminative facial features across spatial dimensions. This approach is partially inspired by Chen et al. \citep{chen2022ost}, who demonstrated that specific facial regions contain important detection cues, but we enhance this concept through our dynamic masking approach.

The LAMM module adaptively recalibrates the attention mask at different network depths, allowing the detector to capture forgery clues across multiple abstraction levels. Similar to how FreqNet \citep{tan2024frequency} incorporates frequency learning within CNNs, our approach integrates facial region awareness within a transformer architecture. We further introduce a region-gated multi-head attention mechanism that selectively modulates attention based on facial regions. This enhances the model's ability to detect subtle inconsistencies that persist across different generation methods.

The main contributions of our work are as follows:
\begin{itemize}
    \item We introduce a region-gated multi-head attention mechanism that selectively modulates attention to key facial areas, enabling the detection of subtle artifacts across different generation methods.
    
    \item We propose a novel facial landmark-guided Vision Transformer architecture with Layer-aware Mask Modulation (LAMM) that dynamically focuses on discriminative facial regions for improved detection of AI-generated face images.
    
    \item We conduct extensive experiments on different datasets generated by various diffusion models and GANs and show that our method significantly outperforms state-of-the-art methods in cross-dataset generalization scenarios.
    
\end{itemize}

%%%%%%%%%%%%%%%%%%%%%%%%%%%%%%%%%%%%%%%%%%%%%%%%%%%%%%%%%%%%%%%%%%%%%%%%

% \section{Typeset section headers in sentence case}

% You presumably are already familiar with the use of \LaTeX. But let 
% us still have a quick look at how to typeset a simple equation: 
% %
% \begin{eqnarray}\label{eq:vcg}
% p_i(\boldsymbol{\hat{v}}) & = &
% \sum_{j \neq i} \hat{v}_j(f(\boldsymbol{\hat{v}}_{-i})) - 
% \sum_{j \neq i} \hat{v}_j(f(\boldsymbol{\hat{v}})) 
% \end{eqnarray}
% %
% Use the usual combination of \verb|\label{}| and \verb|\ref{}| to 
% refer to numbered equations, such as Equation~(\ref{eq:vcg}). 
% Next, a theorem: 

% \begin{theorem}[Fermat, 1637]\label{thm:fermat}
% No triple $(a,b,c)$ of natural numbers satisfies the equation 
% $a^n + b^n = c^n$ for any natural number $n > 2$.
% \end{theorem}

% \begin{proof}
% A full proof can be found in the supplementary material.
% \end{proof}

% Table captions should be centred \emph{above} the table, while figure 
% captions should be centred \emph{below} the figure.\footnote{Footnotes
% should be placed \emph{after} punctuation marks (such as full stops).}
 
% \begin{table}[h]
% \caption{Locations of selected conference editions.}
% \centering
% \begin{tabular}{ll@{\hspace{8mm}}ll} 
% \toprule
% AISB-1980 & Amsterdam & ECAI-1990 & Stockholm \\
% ECAI-2000 & Berlin & ECAI-2010 & Lisbon \\
% ECAI-2020 & \multicolumn{3}{l}{Santiago de Compostela (online)} \\
% \bottomrule
% \end{tabular}
% \end{table}

%%%%%%%%%%%%%%%%%%%%%%%%%%%%%%%%%%%%%%%%%%%%%%%%%%%%%%%%%%%%%%%%%%%%%%%%

\section{Related Work}
In this section, we present a comprehensive overview of existing research on AI-generated face detection. We categorize current techniques into three main categories: image-based detection approaches, frequency-domain detection methods, and attention-guided detection mechanisms.

\subsection{Image-based AI-Generated Face Detection}
Early methods for detecting AI-generated faces primarily focused on exploiting spatial artifacts in the pixel domain. Rossler et al.\citep{rossler2019faceforensics++} utilized the Xception architecture for deepfake detection, demonstrating effective performance on high-quality synthetic media. Several approaches have targeted specific facial regions - Li et al.\citep{li2018ictu} focused on eye region inconsistencies, while Haliassos et al.\citep{haliassos2021lips} examined mouth movement irregularities. Face X-ray\citep{li2020face} identified blending boundaries between forged faces and backgrounds, while SBIs~\citep{shiohara2022detecting} expanded on blending-based forgery detection.

\begin{figure*}[ht]
    \centering

    \includegraphics[width=1.0\textwidth]{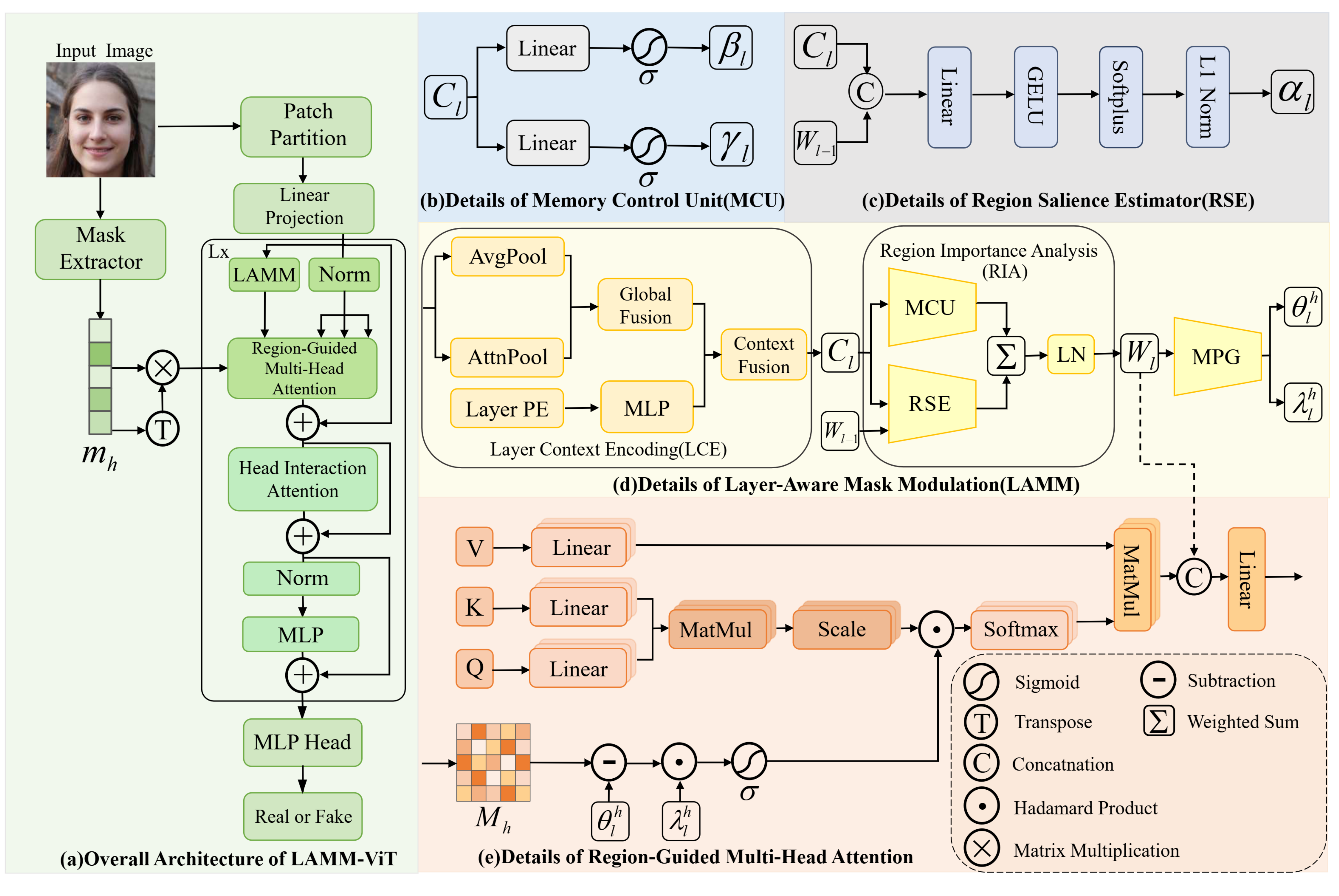} 
    \caption{Overall architecture of the proposed LAMM-ViT. (a) Main pipeline showing the integration of Mask Extractor, Region-Guided Multi-Head Attention, Head Interaction Attention (which is a simple self-attention mechanism), and LAMM within the ViT blocks. (b) Details of the Memory Control Unit (MCU) used in LAMM. (c) Details of the Region Salience Estimator (RSE) used within LAMM's RIA component. (d) Detailed breakdown of the Layer-Aware Mask Modulation (LAMM) module, including Layer Context Encoding (LCE) and Region Importance Analysis (RIA) which generates layer-specific mask weights $W_l$ and gating parameters ($\theta_l^h, \lambda_l^h$) via the Mask Parameter Generator (MPG, detailed in Section~\ref{subsubsec:mpg}). (e) Details of the Region-Guided Multi-Head Attention (RG-MHA) mechanism, showcasing the region gating process.}
    \label{fig:architecture}
\end{figure*}

To address generalization challenges, Wang et al.\citep{wang2020cnn} showed that a detector trained with careful data augmentation on a single specific CNN generator could generalize to unseen architectures. As synthetic media technology advances, approaches like FakeSpotter~\citep{wang2019fakespotter} utilize layer-wise neuron behavior for classification, while Gram-Net~\citep{liu2020global} leverages the Gram matrix to extract global texture as a robust representation. ICT~\citep{dong2022protecting} models identity differences in inner and outer facial regions to better identify inconsistencies across various generation techniques.

\subsection{Frequency-based Detection Methods}
A significant body of research has demonstrated that frequency domain analysis can reveal artifacts invisibly embedded in AI-generated images. Frank et al.\citep{frank2020leveraging} and Durall et al.\citep{durall2020watch} observed that CNN-generated images consistently fail to reproduce realistic spectral distributions, suggesting fundamental limitations in current generative models. F3-Net~\citep{qian2020thinking} explores frequency statistics differences between real and fake images, employing both frequency-aware decomposed components and local frequency statistics to capture forgery patterns.

More specialized frequency-based approaches include FrePGAN~\citep{jeong2022frepgan}, which converts RGB images to frequency maps to highlight generation artifacts, and FDFL~\citep{li2021frequency}, which proposes adaptive frequency feature learning to mine subtle artifacts. LOG~\citep{masi2020two} integrates information from both color and frequency domains through a two-branch recurrent network, while BiHPF~\citep{jeong2022bihpf} emphasizes amplifying artifact magnitudes through dual high-pass filters. Wang et al.~\citep{wang2023dynamic} introduces dynamic graph learning to exploit relation-aware features across spatial and frequency domains.

\subsection{Attention-Guided Detection Approaches}
Attention mechanisms have proven highly effective for deepfake detection by focusing on discriminative facial regions. MAT~\citep{zhao2021multi} pioneered attentional mechanisms to highlight suspicious regions, while Stehouwer et al.~\citep{dang2020detection} introduced attention guided by ground truth manipulation masks. Several approaches have employed multi-headed attention modules to correlate low-level textural features with high-level semantics at different facial regions, though fixed attention paradigms limit adaptation to diverse forgery types.

Recent Vision Transformer (ViT) based approaches show particular promise due to their inherent attention mechanisms. FTCN~\citep{zheng2021exploring} extracts temporal information using specialized attention, while PCL~\citep{zhao2021learning} employs region-specific attention to extract distinct source features. However, Chen et al.~\citep{chen2022self} note that most existing methods use fixed attention weights across network layers, limiting their ability to detect hierarchical facial forgery artifacts. Our LAMM-ViT differs by dynamically adjusting attention at different network depths using facial landmarks and layer-specific parameters, enabling detection of structural inconsistencies across diverse generation techniques at multiple feature abstraction levels.

\section{Methodology}
\label{sec:methodology}

To address the challenge of detecting AI-generated faces with high generalization capability, we propose a novel framework, the Mask-Guided Vision Transformer with Layer-aware Mask Modulation (LAMM-ViT). Our approach enhances the standard Vision Transformer (ViT) \citep{dosovitskiy2020image} architecture by incorporating explicit facial region guidance and dynamically adapting this guidance across different network layers, indexed by $l$. The overall architecture is depicted in Figure~\ref{fig:architecture}(a).

\subsection{Input Processing and Mask Generation}
\label{subsec:input_mask}

Given an input face image $I \in \mathbb{R}^{H \times W \times C}$, where $H$, $W$ are height and width, and $C$ is the number of channels, we first extract facial landmarks using an off-the-shelf detector. We then generate continuous Gaussian masks for $K$ key facial regions (eyes, nose, mouth, etc.), resulting in a multi-channel region mask tensor $R \in \mathbb{R}^{K \times H \times W}$.

Concurrently, the input image $I$ is processed into patch embeddings $X_p \in \mathbb{R}^{N_p \times D}$, where $N_p = HW/P^2$ is the number of patches, $P$ is the patch size, and $D$ is the embedding dimension. A learnable class token $x_{cls}$ is prepended, and positional embeddings $E_{pos}$ are added to form the initial sequence $X_0 \in \mathbb{R}^{(N_p+1) \times D}$.

The region map is then processed by the \textbf{Mask Processor}. This component projects each region mask from size $H \times W$ to a patch-level vector of dimension $N_p$, resulting in $K$ initial mask vectors. To capture relationships between regions, additional combined mask vectors are created via learnable weighted sums of these initial vectors, targeting specific facial groupings. This results in a total of $H$ mask vectors (where $H$ equals the number of attention heads), comprising both individual regions and their combinations.

Finally, a zero vector is prepended to represent the mask for the CLS token, forming the final mask tensor $\mathcal{M} \in \mathbb{R}^{H \times (N_p+1)}$. Each mask vector $m^h \in \mathbb{R}^{N_p+1}$ in $\mathcal{M}$ corresponds to a specific region or combination for potential guidance.

\subsection{Region-Guided Transformer Block}
\label{subsec:transformer_block}

Each Transformer block in our framework enhances standard attention mechanisms by incorporating region-aware processing. This design guides the model to focus on specific facial regions and their interactions, enabling more effective detection of inconsistencies in AI-generated faces.

% \subsubsection{Region-Gated Multi-Head Attention}
% \label{subsubsec:rgmha}

The RG-MHA mechanism, shown in Figure~\ref{fig:architecture}(e), adapts the standard multi-head self-attention to focus on facial region inconsistencies. From the input sequence $X_{l-1}$ of layer $l$, queries $Q_l$, keys $K_l$, and values $V_l$ are computed linearly.

To enable region-aware attention, we first construct an attention gating mask $M^h \in \mathbb{R}^{(N_p+1) \times (N_p+1)}$ for each attention head $h$. This mask is derived from the corresponding mask vector $m^h \in \mathbb{R}^{N_p+1}$ from the mask tensor $\mathcal{M}$ via an outer product:
\begin{equation}
M^h = m^h (m^h)^T,
\end{equation}
where $m^h$ is the mask vector for head $h$, and $M^h$ is the resulting attention gating mask.

Using this mask, we compute a region gate $G_l^h \in \mathbb{R}^{(N_p+1) \times (N_p+1)}$ that selectively emphasizes attention to specific facial regions and their interactions:
\begin{equation}
G_l^h = \sigma(\lambda_l^h \cdot (M^h - \theta_l^h)),
\end{equation}
where $\sigma$ is the sigmoid function, and $\lambda_l^h $ and $\theta_l^h $ are layer-specific gating parameters dynamically generated by the Mask Parameter Generator (MPG) in Section \ref{subsubsec:mpg}.

The region gate modulates the attention mechanism by element-wise multiplication with the attention scores before softmax normalization:
\begin{equation}
\text{Attention}(Q_l^h, K_l^h, V_l^h) = \text{softmax}\left(\frac{Q_l^h (K_l^h)^T}{\sqrt{d}} \odot G_l^h\right)V_l^h,
\end{equation}
where $\odot$ is element-wise multiplication, $d=D/H$ is the dimension per head for a Transformer with $H$ attention heads, and $Q_l^h$, $K_l^h$, and $V_l^h$ are the query, key, and value matrices for head $h$ in layer $l$.

Finally, the outputs from all heads are combined using a weighted concatenation strategy. Each head's output is weighted by its corresponding layer-specific weight $W_l^h$, which is an element of the mask weight vector $W_l = [W_l^1, W_l^2, ..., W_l^H]$ generated by the LAMM module for layer $l$, where $h \in \{1,2,...,H\}$ represents the index of each attention head out of a total of $H$ attention heads.

After processing through the attention mechanism and feed-forward network, the output of each layer forms the sequence $X_l$, which serves as input to the subsequent layer.

\subsection{Layer-Aware Mask Modulation}
\label{subsec:lamm}

The LAMM module dynamically adjusts how different facial regions influence attention at each network depth. This enables the model to progressively refine its focus on discriminative facial features across different abstraction levels. As illustrated in Figure~\ref{fig:architecture}(d), LAMM generates layer-specific parameters that control the RG-MHA mechanism: mask weights $W_l$ for weighting head outputs, and gating parameters that control the strength and threshold of regional attention.

\subsubsection{Layer Context Encoding}
\label{subsubsec:lce}

The Layer Context Encoding component captures the network's state at each layer $l$, providing essential context for adaptive parameter generation. LCE computes a context vector $C_l$ specific to encoder layer $l$ by combining information from two sources.

First, layer position information $PE_l$ is encoded for each layer index $l$ using a learned embedding approach. This embedding captures the depth-specific characteristics of the network.

Second, global features $g_l$ are extracted from the input sequence through pooling methods and then combined:
\begin{equation}
g_l = \text{LayerNorm}(\text{Linear}(\text{Concat}(g_l^{avg}, g_l^{att}))),
\end{equation}
where $g_l^{avg}$ and $g_l^{att}$ are the average-pooled and attention-pooled features.

Finally, the context fusion mechanism integrates the layer position information with the global feature state:
\begin{equation}
C_l = \text{LayerNorm}(\text{MLP}(\text{Concat}(g_l, PE_l))),
\end{equation}

\subsubsection{Region Importance Analysis}
\label{subsubsec:ria}

The Region Importance Analysis component determines which facial regions should receive more attention at each layer. RIA dynamically updates the mask weights $W_l$ used in RG-MHA by leveraging both current layer information and accumulated knowledge from previous layers. It consists of a Region Salience Estimator (RSE) that assesses the current importance of each region, and a Memory Control Unit (MCU) that balances new information with previously learned patterns.

These components work together to update the mask weights through a recurrent-like mechanism:
\begin{equation}
W_l = \gamma_l \odot \alpha_l + \beta_l \odot W_{l-1},
\end{equation}
where $\alpha_l$ represents the new region importance scores computed by the RSE based on layer context $C_l$, $\gamma_l$ and $\beta_l$ are adaptive coefficients produced by the MCU that control the balance between new information and historical knowledge.

\subsubsection{Mask Parameter Generator (MPG)}
\label{subsubsec:mpg}

The Mask Parameter Generator produces the specific parameters that control regional gating in the attention mechanism. Taking the layer context $C_l$ and the mask weights $W_l$ as input, it generates two sets of parameters.

The gating strength parameters $\lambda_l$ control how strongly each head emphasizes its assigned regions:
\begin{equation}
\lambda_l = \text{FC}(C_l) \odot \text{MLP}(\text{Concat}(C_l, W_l)),
\end{equation}
where $\text{FC}$ is a fully connected layer and $\text{MLP}$ is a multi-layer perceptron. The parameter $\lambda_l$ is then used to derive head-specific gating strength parameters $\lambda_l^h$ for each attention head.

The gating threshold parameters $\theta_l$ determine the sensitivity of each head to its assigned regions:
\begin{equation}
\theta_l = \theta_{base} + \text{MLP}(\text{Concat}(C_l, W_l)),
\end{equation}
where $\theta_{base}$ is a base threshold value. Similarly, $\theta_l$ is used to derive head-specific threshold parameters $\theta_l^h$. These adaptively generated parameters enable each attention head to dynamically adjust its regional focus based on both network depth and learned feature representations.

\subsection{Loss Function}
\label{subsec:loss}

Our training objective combines classification accuracy with a novel diversity-promoting mechanism specifically designed to enhance generalization across various generation techniques.

For the primary task of distinguishing between real and AI-generated faces, we employ the standard Cross-Entropy loss ($\mathcal{L}_{\text{ce}}$):
\begin{equation}
\mathcal{L}_{\text{ce}} = -\frac{1}{N}\sum_{i=1}^{N}[y_i\log(\hat{y}_i) + (1-y_i)\log(1-\hat{y}_i)],
\end{equation}
where $y_i$ is the ground truth label, $\hat{y}_i$ is the predicted probability for each sample, and $N$ is the number of samples.

To address the challenge of detecting diverse forgery patterns across different generation techniques, we introduce a novel Mask Diversity Loss ($\mathcal{L}_{\text{div}}$). This component leverages our LAMM-ViT's region-guided attention mechanism to encourage the model to utilize different facial region combinations when analyzing different samples. The key insight is that various generative techniques produce artifacts in different facial regions, requiring multiple detection strategies tailored to different artifact patterns.

For this purpose, we utilize the layer mask weights $W_l$ for each input sample $i$ at layer $l$. These weights reflect how the model assigns region importance during processing.

We first define the cosine similarity between mask weight vectors for two samples:
\begin{equation}
\cos(W_{l,i}, W_{l,j}) = \frac{W_{l,i} \cdot W_{l,j}}{||W_{l,i}|| \cdot ||W_{l,j}||},
\end{equation}
where $W_{l,i} \cdot W_{l,j}$ is the dot product between vectors, and $||W||$ represents the Euclidean norm.

The Mask Diversity Loss measures the average similarity between all pairs of sample mask weights across all network layers:
\begin{equation}
\mathcal{L}_{\text{div}} = \frac{1}{L} \sum_{l=1}^{L} \frac{\sum_{i=1}^{N}\sum_{j=1, j \neq i}^{N} \cos(W_{l,i}, W_{l,j})}{N(N-1)},
\end{equation}
where $L$ is the total number of layers in the network. Higher pairwise similarity indicates lower diversity in attention strategies, which is penalized by this loss term.

Our total loss function combines these components:
\begin{equation}
\mathcal{L} = \mathcal{L}_{\text{ce}} + \eta \mathcal{L}_{\text{div}},
\end{equation}
where $\eta = 0.2$ was found to yield the best results in our experiments. This balanced approach ensures accurate classification while simultaneously promoting diverse and adaptive attention mechanisms across samples, enhancing generalization capabilities for detecting various types of generation artifacts.

\section{Experiments}
In this section, we first introduce the overall experimental setup, and then present extensive experimental results to demonstrate the superiority of our method.

\begin{table*}[!t]
\centering
\caption{Performance comparison (ACC/AP \%) with state-of-the-art methods across 18 diverse AI-generated face models (GANs and Diffusion). Our method (LAMM-ViT) demonstrates superior generalization, achieving the highest mean ACC (94.09\%) and AP (98.62\%). Best results per generator are highlighted in bold.}
\resizebox{\textwidth}{!}{% Make table fit width if needed
\begin{tabular}{lccccccc} % Changed alignment for better readability
\toprule
Generator & Wang \citep{wang2020cnn} & F3Net \citep{qian2020thinking} & Grag \citep{gragnaniello2021gan} & LGrad \citep{tan2023learning} & Ojha \citep{ojha2023towards} & FreqNet \citep{tan2024frequency} & Ours (LAMM) \\
        \midrule
        AttGAN & 81.81/98.98 & 84.72/97.93 & 99.63/99.99 & 99.92/100.0 & 84.26/94.33 & 99.54/99.97 & \textbf{82.76}/\textbf{95.70} \\
        MMDGAN & 92.50/98.32 & 99.50/\textbf{100.0} & 99.50/99.99 & \textbf{100.0}/\textbf{100.0} & 87.75/95.69 & \textbf{100.0}/\textbf{100.0} & 99.75/\textbf{100.0} \\
        MSGGAN & 92.50/99.90 & \textbf{100.0}/\textbf{100.0} & 99.50/\textbf{100.0} & \textbf{100.0}/\textbf{100.0} & 77.50/90.83 & 99.25/\textbf{100.0} & 97.00/99.96 \\
        StarGAN & 72.83/82.94 & 56.11/83.24 & 99.82/\textbf{100.0} & 99.87/99.99 & 95.44/98.96 & 99.25/99.98 & \textbf{79.12}/\textbf{94.63} \\
        STGAN & 95.75/99.32 & \textbf{100.0}/\textbf{100.0} & 99.25/99.87 & \textbf{100.0}/\textbf{100.0} & 83.75/93.71 & 99.75/\textbf{100.0} & 97.75/99.78 \\
        StyleGAN & 92.76/98.60 & 87.58/94.99 & 49.96/77.99 & 50.74/88.87 & 88.46/95.88 & 50.11/55.22 & \textbf{97.40}/\textbf{99.73} \\
        StyleGAN2 & 92.55/98.43 & 84.72/94.98 & 49.89/63.27 & 51.54/82.79 & 85.38/94.33 & 50.05/51.02 & \textbf{97.14}/\textbf{99.65} \\
        StyleGAN3 & 94.06/99.00 & \textbf{99.99}/\textbf{100.0} & 99.70/\textbf{100.0} & 99.95/\textbf{100.0} & 96.17/\textbf{100.0} & 99.85/\textbf{100.0} & 97.19/99.54 \\
        VQGAN & 86.76/94.29 & 50.30/83.45 & 49.77/55.80 & 50.72/77.73 & 81.60/91.97 & 52.95/79.20 & \textbf{94.56}/\textbf{98.68} \\
        ProGAN & 93.50/99.11 & \textbf{99.74}/\textbf{100.0} & 99.66/\textbf{100.0} & \textbf{99.95}/\textbf{100.0} & 93.06/97.88 & 99.46/\textbf{100.0} & 96.62/99.51 \\
        Midjourney & 92.50/98.13 & \textbf{100.0}/\textbf{100.0} & \textbf{100.0}/\textbf{100.0} & \textbf{100.0}/\textbf{100.0} & \textbf{100.0}/\textbf{100.0} & \textbf{100.0}/\textbf{100.0} & 95.00/98.68 \\
        IF & 90.59/98.66 & \textbf{100.0}/\textbf{100.0} & 99.01/\textbf{100.0} & \textbf{100.0}/\textbf{100.0} & 96.53/\textbf{100.0} & 99.01/\textbf{100.0} & 97.03/99.47 \\
        DALLE2 & 80.49/91.36 & \textbf{100.0}/\textbf{100.0} & \textbf{100.0}/\textbf{100.0} & \textbf{100.0}/\textbf{100.0} & 97.56/\textbf{100.0} & \textbf{100.0}/\textbf{100.0} & 91.46/97.83 \\
        DCFACE & 81.43/90.29 & 50.00/81.05 & 53.34/76.54 & 49.93/40.57 & 72.05/85.54 & 49.97/46.00 & \textbf{97.25}/\textbf{99.33} \\
        Latent Diffusion & 93.95/99.47 & \textbf{100.0}/\textbf{100.0} & 99.79/\textbf{100.0} & 99.94/\textbf{100.0} & 96.13/\textbf{100.0} & 99.88/\textbf{100.0} & 97.38/99.69 \\
        Palette & 84.67/92.76 & 50.00/76.66 & 49.79/38.61 & 50.88/77.96 & 51.88/56.31 & 53.75/79.75 & \textbf{93.29}/\textbf{98.28} \\
        SD Inpainting & 84.83/92.59 & \textbf{99.54}/\textbf{100.0} & 99.70/\textbf{100.0} & \textbf{99.92}/\textbf{100.0} & 95.58/99.51 & 98.98/\textbf{100.0} & 88.98/96.52 \\
        SD v1.5 & 92.05/97.45 & \textbf{99.99}/\textbf{100.0} & 99.65/\textbf{100.0} & \textbf{99.89}/\textbf{100.0} & 95.74/99.53 & 98.62/\textbf{100.0} & 93.89/98.20 \\
        \midrule
        Mean & 88.64/95.53 & 86.79/95.12 & 86.00/89.56 & 86.29/92.66 & 87.71/94.13 & 86.13/89.50 & \textbf{94.09}/\textbf{98.62} \\
\bottomrule
\end{tabular}%
}

\label{tab:sota_comparison}
\end{table*}

\subsection{Experimental Setup}
\textbf{Datasets.} We conduct experiments on a subset of the AI-Face-FairnessBench~\citep{lin2024ai} dataset, which includes both real and AI-generated face images. The real images are sourced from IMDB-WIKI datasets. For AI-generated images, we include a diverse collection from multiple generative models, categorized into GANs and Diffusion Models (DMs). The GAN-based models include AttGAN, MMDGAN, StarGAN, MSGGAN, STGAN, StyleGAN, StyleGAN2, StyleGAN3, VQGAN and ProGAN. The diffusion-based models include DALLE2, IF, Midjourney, DCFACe, Latent Diffusion, Palette, SD v1.5, and SD Inpainting.

For training our model, we use real images from IMDB-WIKI datasets, along with fake images generated by StyleGAN3, Latent Diffusion, and SD V1.5. We deliberately include multiple generative models in our training set to enable our region-based method to learn the relationship between face regions between different generative techniques which is essential for generalizable detection.

\textbf{Implementation Details.} All images are resized to 224$\times$224 resolution for consistency. We extract facial landmarks from each image using DLIB's\citep{sagonas2016300} 68-point facial landmark detector and generate region-specific masks for the eight facial areas as described in Section 3. We implement our Region-Gated Vision Transformer with 12 transformer layers, 12 attention heads, and embedding dimension of 768. We train using the AdamW optimizer with a learning rate of $1\times10^{-4}$, weight decay of 0.05, and batch size of 64. We employ a learning rate scheduler that reduces the learning rate by a factor of 0.5 when validation performance plateaus for 3 consecutive epochs. All models are trained for a maximum of 100 epochs with early stopping based on validation accuracy. We implement our approach with PyTorch and use mixed-precision training to improve efficiency.

\textbf{Evaluation Metrics.} Following previous works \citep{frank2020leveraging, ojha2023towards, wang2020cnn}, we evaluate performance using Average Precision (AP) and Accuracy (ACC). For computing ACC, we use a classification threshold of 0.5 following standard practice.

\textbf{Baselines.} We compare our method with several state-of-the-art approaches:
(i) Wang et al. \citep{wang2020cnn}, which demonstrated that a standard classifier trained on a single CNN generator with careful data augmentation can generalize surprisingly well to unseen architectures;
(ii) F3Net \citep{qian2020thinking}, which proposed mining frequency-aware clues using decomposed image components and local frequency statistics within a two-stream framework;
(iii) Gragnaniello et al. \citep{gragnaniello2021gan}, which analyzed the generalization ability of detectors across different GAN architectures and challenging scenarios, studying the impact of augmentation and training strategies;
(iv) LGrad \citep{tan2023learning}, which introduced using gradients from a pretrained model as a generalized representation for GAN-generated artifacts;
(v) Ojha et al. \citep{ojha2023towards}, which proposed using features from large pretrained models (like CLIP), not explicitly trained for fake detection, to achieve better generalization across diverse generative model families;
and (vi) FreqNet \citep{tan2024frequency}, which leverages frequency space domain learning for improved generalizability. % Added FreqNet

\subsection{Comparison with State-of-the-Art Methods}
\label{sec:sota_comparison}

To evaluate the effectiveness and generalization capabilities of LAMM-ViT, we perform extensive comparisons with state-of-the-art detection methods across 18 diverse generative models. The quantitative results are summarized in Table~\ref{tab:sota_comparison}.

\textbf{Cross-dataset Performance and Model Generalizability.} Our model shows exceptional cross-model generalization performance, achieving \textbf{94.09\%} mean ACC and \textbf{98.62\%} mean AP across all tested generators, significantly outperforming the strongest baseline (Wang et al.~\citep{wang2020cnn} with 88.64\% ACC and 95.53\% AP) by \textbf{+5.45\%} in ACC and \textbf{+3.09\%} in AP.

The key strength of LAMM-ViT lies in its consistent performance across diverse generator types. While competing methods exhibit extreme performance variations, our approach maintains robust accuracy with no catastrophic failures. For instance, F3Net~\citep{qian2020thinking} achieves perfect accuracy on several generators (MMDGAN, MSGGAN) but drops to chance level on others (VQGAN, DCFACE), and FreqNet~\citep{tan2024frequency} similarly shows inconsistent results across different models.

LAMM-ViT particularly excels on challenging generators where baselines struggle. On StyleGAN and StyleGAN2 where methods like Gragnaniello et al.~\citep{gragnaniello2021gan} achieve only $\sim$50\% accuracy, our approach maintains excellent performance (97.40\% and 97.14\% respectively). For difficult diffusion models like DCFACE and Palette, LAMM-ViT achieves 97.25\% and 93.29\% accuracy where most competitors perform poorly.

Most notably, LAMM-ViT demonstrates balanced effectiveness across both GAN-based and diffusion-based models without favoring the generative family it was trained on—a crucial advantage for real-world deployment where source generators are typically unknown. This consistent performance suggests our method captures fundamental structural inconsistencies common across generation techniques rather than overfitting to specific artifacts.

\begin{figure}[!t]
\centering
\includegraphics[width=\columnwidth]{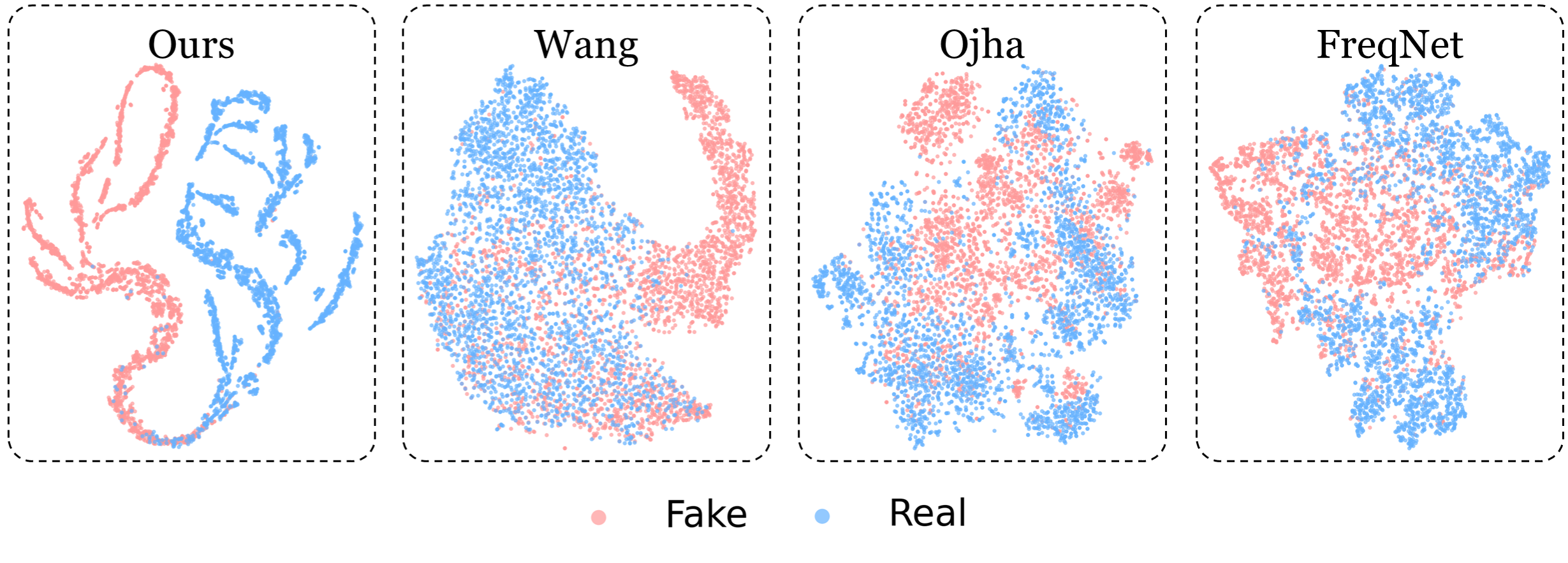}
\caption{The t-SNE visualization of features extracted from our model and some state-of-the-art models~\citep{wang2020cnn, ojha2023towards, tan2024frequency} on all test datasets. Our model demonstrates clearer separation between real and synthetic clusters compared to competing approaches.
}
\label{fig:tsne}
\end{figure}

\textbf{Feature Space Analysis.} The t-SNE visualization in Figure~\ref{fig:tsne} reveals a distinct separation between real images (blue cluster) and various synthetic image clusters. Unlike previous methods~\citep{wang2020cnn,ojha2023towards,tan2024frequency} where real and fake clusters often significantly overlap, our feature space maintains clear decision boundaries with logical positioning of different generator families. This structured representation confirms that LAMM-ViT learns the discriminative features of generalization across generative techniques rather than merely detecting model-specific artifacts that frequency-based methods typically target.

% t-SNE Visualization Section

\subsection{Robustness to Image Perturbations}

\label{sec:robustness}

\begin{figure*}[ht]
\centering
\includegraphics[width=1\textwidth]{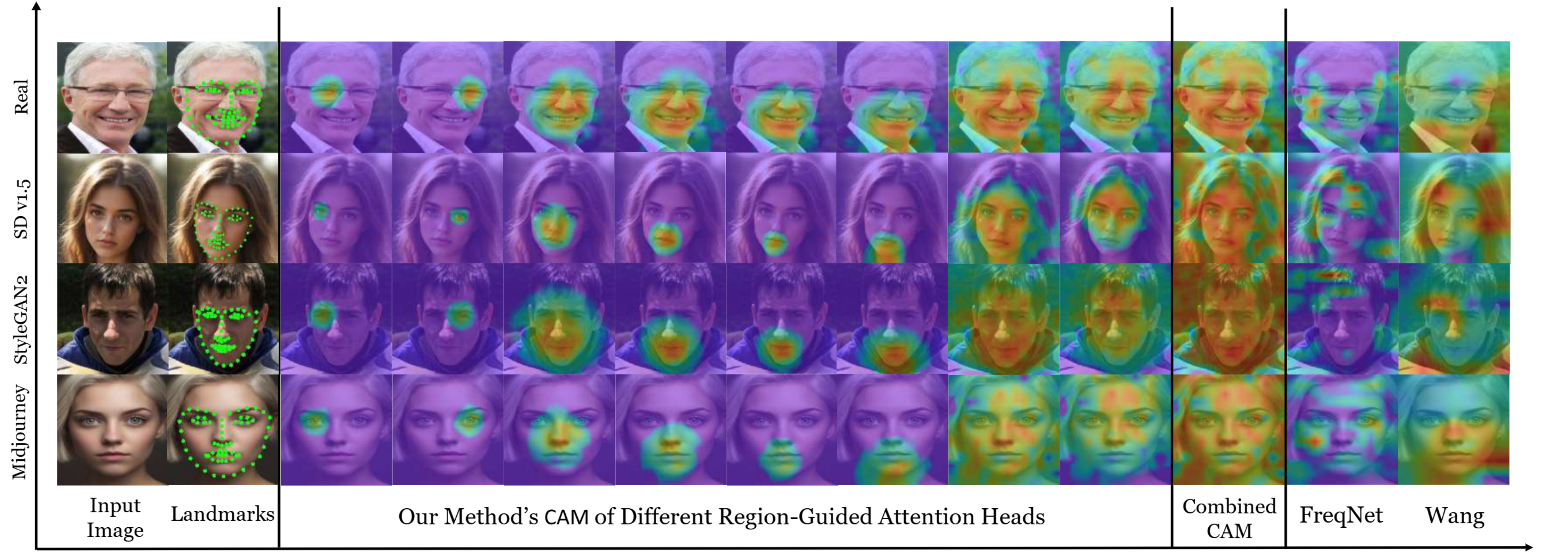} % Make sure this path is correct
\caption{CAM visualizations of LAMM-ViT on various AI-generated faces from Midjourney, StyleGAN2, and SDv1.5. Comparison between our regional head-specific CAM, combined CAM, and baseline methods (FreqNet and Wang). }
\label{fig:cam}
\end{figure*}

\begin{table}[!t]
\vspace{-0.4cm}
\centering
\caption{Robustness evaluation against common image perturbations.}
\renewcommand{\arraystretch}{1.2}  % 控制行间距，可以调整这个值
\begin{tabular}{l|cc|cc|cc}
\hline
\multirow{2}{*}{Perturbed} & \multicolumn{2}{c|}{GAN-based} & \multicolumn{2}{c|}{Diffusion-based} & \multicolumn{2}{c}{Mean} \\
\cline{2-7}
 & ACC & AP & ACC & AP& ACC & AP \\
\hline
No                 & 93.93 & 98.72 & 94.29 & 98.50 & 94.09 & 98.62 \\ 
noise    & 94.48 & 98.77 & 96.25 & 99.14 & 95.27 & 98.93 \\ 
jpeg  & 94.33 & 98.77 & 94.59 & 99.17 & 94.45 & 98.95 \\ 
blur      & 94.62 & 98.91 & 94.51 & 98.57 & 94.57 & 98.76 \\ 
cropping       & 89.14 & 92.36 & 92.71 & 95.84 & 90.73 & 93.91 \\ 
combined   & 89.67 & 93.71 & 91.95 & 95.87 & 90.68 & 94.67 \\ 
\hline
\end{tabular}
\label{tab:robustness}
\end{table}

We evaluated LAMM-ViT's resilience against common image manipulations that typically occur in real-world scenarios. Following Frank et al. \citep{frank2020leveraging}, we applied perturbations to test images with a probability of 50\%, including gaussian noise, jpeg compression, blurring, cropping, and a challenging combined scenario. As shown in Table~\ref{tab:robustness},Our model shows significant stability across most perturbations without retraining. The performance under gaussian noise, jpeg compression and blurring is always high, and high accuracy and precision indicators are maintained in these common distortions. Cropping causes a modest decline since it removes important spatial context, yet performance remains robust. Even under the demanding combined perturbation scenario, LAMM-ViT maintains strong performance with only a manageable drop compared to standard conditions. This consistency across different perturbations emphasizes the advantage of LAMM-ViT in focusing on structural relationships between facial regions rather than low-level textures or frequency artifacts that are easily degraded, highlighting its applicability to powerful real-world deployments of increasingly complex synthetic media.

\subsection{Ablation Study}

We conduct ablation experiments to evaluate the contribution of individual components in our LAMM-ViT framework and validate the effectiveness of our loss function design.

\textbf{Ablation Study on Components.} To understand the impact of each proposed module, we systematically evaluate different architectural configurations by selectively including key components: the initial region Mask guidance, Region-Guided Multi-Head Attention (RG-MHA), and Layer-aware Mask Modulation (LAMM). As shown in Table~\ref{tab:ablation}, the standard Vision Transformer baseline achieves reasonable performance, but simply adding static facial region masks without the corresponding guidance mechanisms leads to significant performance degradation. This suggests that static masks alone inappropriately restrict the ViT's attention patterns. Similarly, including RG-MHA or LAMM alone yields suboptimal results, indicating the synergistic nature of these components rather than independence. When all components are integrated, our full model demonstrates substantially superior performance, confirming that the dynamic relationship between region-directed attention and layer-specific modulation is critical for effectively capturing forgery patterns.

\textbf{Ablation Study on Loss Functions.} Table~\ref{tab:ablation_loss} presents the comparison between training with only Cross-Entropy loss (CE), only our proposed Diversity loss, and the combined loss function. The results demonstrate that while Cross-Entropy loss alone provides reasonable classification performance, it cannot match the combined approach. Training with only the Diversity loss predictably fails to provide sufficient classification guidance. However, when both losses are combined, we observe substantial improvement, confirming that our proposed Diversity loss successfully encourages the model to learn multiple detection strategies targeting different facial regions, enhancing generalization across diverse generation techniques.

\begin{table}[!t]
\vspace{-0.41cm} 
\centering
\caption{Ablation study on LAMM-ViT components. Each experiment follows the same configuration as our main experiments, varying only the inclusion of specific architectural components.}
\vspace{0.05cm}  % 在这里添加表格标题与表格内容之间的距离
\renewcommand{\arraystretch}{1.2}  % 将行间距增加到原来的1.2倍
\begin{tabular}{c ccc|cc}
\hline
\multicolumn{4}{c|}{Component} & \multicolumn{2}{c}{Mean} \\
\cline{1-4} \cline{5-6}
ViT & Mask & MGMHA & LAMM & ACC & AP \\
\hline
\checkmark & - & - & - & 82.37 & 89.54 \\ % Baseline ViT (All blank)
\checkmark & \checkmark & - & - & 52.57 & 53.83 \\ % ViT + Mask
\checkmark & \checkmark & \checkmark & - & 56.62 & 61.91 \\ % ViT + Mask + MGMHA (+HeadInteract)
\checkmark & \checkmark & - & \checkmark & 43.82 & 45.75 \\ % ViT + Mask + LAMM
\checkmark & \checkmark & \checkmark & \checkmark & \textbf{94.09} & \textbf{98.62} \\ % Full Model
\hline
\end{tabular}
\label{tab:ablation}
\vspace{-0.05cm} 
\end{table}

\begin{table}[t]
\vspace{-0.41cm} 
\centering
\caption{Ablation study on loss function configurations. All experiments use identical settings to our main experiments, only changing the loss function components.}
\vspace{0.05cm}  % 在这里添加表格标题与表格内容之间的距离
\renewcommand{\arraystretch}{1.2}
\begin{tabular}{cc|cc}
\hline
\multicolumn{2}{c|}{Loss} & \multicolumn{2}{c}{Mean} \\
\cline{1-2} \cline{3-4}
Lce & Ldiv & ACC & AP \\
\hline
\checkmark & \checkmark & \textbf{94.09} & \textbf{98.62} \\
\checkmark & - & 89.97 & 95.73 \\
- & \checkmark & 49.95 & 51.21 \\
\hline
\end{tabular}
\label{tab:ablation_loss}
\end{table}

\subsection{Visualization of Region-Gated Attention}

\label{sec:visualization}

To further evaluate the interpretability and validity of our region-gated attention mechanism, we visualized the attention patterns of facial images. We apply Grad-CAM~\citep{selvaraju2017grad} for representation visualization. As shown in Figure \ref{fig:cam}, our approach offers several noteworthy insights. Firstly, it becomes evident that our method excels in extracting diverse spatial attention cues through its specialized attention heads. Regional CAM visualizations show that the different attention in the LAMM-ViT is focused on different facial regions with minimal overlap which demonstrates the effectiveness of our region-guided design. This provides strong evidence of the orthogonality within extracted regional representations. Our method also captures transition zones between different parts of the face, revealing important spatial relationships. In contrast, baseline methods (FreqNet and Wang et al.) demonstrate more scattered focus, often with attention concentrated on limited or less semantically meaningful areas. These visualizations confirm that our region-gated attention mechanism effectively guides the model to recognize multiple face regions independently thereby contributing to LAMM-ViT robust cross-model generation performance.

\section{Conclusion}
\label{sec:conclusion}
In this paper, we presented LAMM-ViT, a novel Vision Transformer architecture for detecting AI-generated faces with robust cross-model generalization. By integrating Region-Guided Multi-Head Attention with Layer-aware Mask Modulation, our approach focuses on structural inconsistencies between facial regions—a common weakness across generation techniques. Experiments demonstrated LAMM-ViT's superior performance, achieving 94.09\% mean accuracy and 98.62\% mean AP across 18 different generative models, significantly outperforming state-of-the-art methods. The model maintains consistent performance on both GAN-based and diffusion-based generators without catastrophic failure, highlighting its practical utility against evolving synthetic media threats. Our approach shows that region-focused, hierarchical attention mechanisms offer a promising direction for developing generalizable forgery detection systems capable of addressing increasingly photorealistic AI-generated content.

% \newpage

% \section{Citations and references}

% Include full bibliographic information for everything you cite, 
% be it a book \citep{pearl2009causality}, a journal article 
% \citep{grosz1996collaborative,rumelhart1986learning,turing1950computing}, 
% a conference paper \citep{kautz1992planning}, or a preprint 
% \citep{perelman2002entropy}. The citations in the previous sentence are 
% known as \emph{parenthetical} citations, while this reference to the 
% work of \citet{turing1950computing} is an \emph{in-text} citation.
% The use of \BibTeX\ is highly recommended. 

%%%%%%%%%%%%%%%%%%%%%%%%%%%%%%%%%%%%%%%%%%%%%%%%%%%%%%%%%%%%%%%%%%%%%%%%

%%% Use this environment to include acknowledgements (optional).
%%% This will be omitted in doubleblind mode.

% \begin{ack}
% By using the \texttt{ack} environment to insert your (optional) 
% acknowledgements, you can ensure that the text is suppressed whenever 
% you use the \texttt{doubleblind} option. In the final version, 
% acknowledgements may be included on the extra page intended for references.
% \end{ack}

%%%%%%%%%%%%%%%%%%%%%%%%%%%%%%%%%%%%%%%%%%%%%%%%%%%%%%%%%%%%%%%%%%%%%%%%

%%% Use this command to include your bibliography file.

% \bibliographystyle{ieeetr}
\bibliography{mybibfile} % 引用文件名

\end{document}